\documentclass{article}

\usepackage{microtype}
\usepackage{graphicx}
\usepackage{subfigure}
\usepackage{booktabs} 
\usepackage{multirow}
\usepackage{array} 
\usepackage{makecell}
\newcolumntype{C}[1]{>{\centering\arraybackslash}p{#1}}

\usepackage{hyperref}

\usepackage[accepted]{icml2024}

\usepackage{amsmath}
\usepackage{amssymb}
\usepackage{mathtools}
\usepackage{amsthm}
\usepackage{tabularx}
\usepackage{multirow}

\usepackage[capitalize,noabbrev]{cleveref}

\theoremstyle{plain}

\theoremstyle{definition}

\theoremstyle{remark}

\usepackage[textsize=tiny]{todonotes}

\icmltitlerunning{Investigating Relational Reasoning in VLMs}
\usepackage[utf8]{inputenc}

\begin{document}

\twocolumn[
\icmltitle{Investigating Relational Reasoning in VLMs}



\icmlsetsymbol{equal}{*}

\begin{icmlauthorlist}
\icmlauthor{Adhithya Laxman Ravi Shankar Geetha}{equal,yyy}
\icmlauthor{Aulia Kharis Rakhmasari}{equal,yyy}
\icmlauthor{Haleema Ramzan}{equal,yyy}
\icmlauthor{Xander Yap}{equal,yyy}
\end{icmlauthorlist}

\icmlaffiliation{yyy}{Department of Computer Science, ETH Zurich, Switzerland}

\icmlcorrespondingauthor{
Adhithya Laxman Ravi Shankar Geetha,
Aulia Rakhmasari,
Haleema Ramzan,
Xander Yap
}{%
aravishankar@ethz.ch,
arakhmasari@ethz.ch,
hramzan@ethz.ch,
xanyap@ethz.ch. Our code is available at: \href{https://github.com/xandaaaa/Investigating-relational-reasoning-in-VLMs}{Github}
}

\icmlkeywords{Machine Learning, Deep Learning}

\vskip 0.3in
]



\printAffiliationsAndNotice{\icmlEqualContribution} 
\begin{abstract}
Vision-Language Models (VLMs) achieve strong performance in visual reasoning tasks, but it remains unclear whether they understand visual relations, or simply employ shortcuts such as language cues or priors. To investigate this, we use the Qwen3-VL-4B  \cite{bai2025qwen3vltechnicalreport}, a modern VLM, to decode how visual information is encoded across depths. For this, we propose a synthetic dataset of simple geometric shapes for controlled analysis, along with queries crafted to precisely test language cues. Furthermore, the dataset is modified to test causal reliance on visual evidence. Our results show that current VLMs combine genuine visual reasoning with shortcut strategies primarily rooted in language cues.

\end{abstract}

\section{Introduction}
Visual data contains primitives (what exists), their attributes, and the relationships among them. These fundamental properties can be used to describe almost any visual information, with complexity added as a result of a combination of them. While seemingly straight-forward for humans, it encodes rich and complex implicit details that cannot directly be mapped to an explicit domain. For instance, an object A present at the left of object B in an image does not contain any pixel demarcations for this relationship, but rather an understanding of relative space on the perceivers' end.

With the recent rise of large-scale Vision--Language Models (VLMs) such as LLaVA \cite{liu2024improvedbaselinesvisualinstruction}, GPT-4V \cite{wu2024gpt4visionhumanalignedevaluatortextto3d}, and Gemini \cite{geminiteam2025geminifamilyhighlycapable}, that seem to achieve strong performance on many multi-modal tasks, the question remains as to how the visual information is internally \emph{understood} by such models. A thorough comprehension would be marked by understanding beyond the explicit representation. Recent works have explored the performance of such models on diagrammatic information \cite{lu2023mathvista, zhang2024mathverse}, where the work by \cite{hou2024vision} shows that VLMs often struggle with \emph{relational} reasoning beyond entity recognition (e.g., compositional reasoning on diagram reasoning benchmarks) \cite{johnson2017clevr, masry2022chartqa}.

Motivated by concerns that some apparent success can come from shortcuts rather than grounded visual--relational understanding, our work aims to study how visual information is represented \emph{inside} a VLM, and how sensitive predictions are to modifying the visual evidence \cite{hou2024vision} and language cues. Our work has the following key findings: \textbf{(i)} VLMs encode different visual information at different depths, \textbf{(ii)} Recognition tasks are understood on an architectural level, but relational tasks do not exhibit any consistent pattern, and \textbf{(iii)} VLMs show heavy reliance on presence of explicit visual elements. Our findings confirm previous works, and provide a deeper understanding of information processing within VLMs at an architectural level, opening avenues for advanced architectural modifications.

\begin{table*}[h!]
\begin{tabularx}{\textwidth}{p{0.15\textwidth}p{0.15\textwidth}X}
    \textbf{Analysis Target} & \textbf{Query Type} & \textbf{Example}  \\
    \midrule
    \multirow{4}{*}{\textbf{Recognition}}
 & Count & How many shapes are in this image? \\
 & Shape & Does this image have a square? \\
 & Color & Does this image have a green shape? \\
 & Shape \& Color & Does this image have a green square? \\
 \midrule
    \multirow{3}{*}{\textbf{Relational}}
& Arrow Existence & Which objects are connected? \\
 & Arrow Direction & Where does the arrow between the cyan circle and orange rectangle point to? \\
 & Implicit Position & What is the position of the orange rectangle with respect to the cyan circle? \\
    \bottomrule
\end{tabularx}
\vskip -0.1in
\caption{An overview of the variations of query types and their intended analysis target. Examples are provided from the dataset.}
\label{tab:queries}
\end{table*}

\section{Related Work}
Mechanistic interpretability work has shown that transformer attention patterns can encode structured dependencies and can be analyzed to study syntax and compositional reasoning in language models. In VLMs, attention-based analysis has more commonly been used for object localization and text--image grounding than for diagnosing whether attention captures \emph{inter-entity relations} in a mechanistic sense \cite{clark2019does, elhage2021mathematical}.

For evaluation of such tasks, relational reasoning benchmarks such as GQA \cite{ainslie2023gqa} and CLEVR \cite{johnson2017clevr} highlight systematic weaknesses in relation-centric querying, while diagram benchmarks such as ChartQA \cite{masry2022chartqa} provide complementary tests requiring visual and logical reasoning over structured graphics. Recent work by \cite{hou2024vision} argues that VLMs may rely on shortcuts rather than true relational understanding, motivating controlled tests and internal diagnostics. Building on this, we combined controlled relational tasks with attention-based probing to understand how relational information is encoded in VLMs.

\section{Methodology}

\subsection{Synthetic Dataset Creation}
\label{dataset-creation}
\begin{figure}[h]
\centering

    \centering
    \includegraphics[height=3.5cm, keepaspectratio]{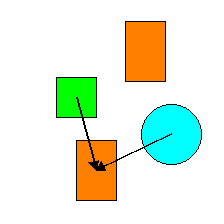}
    \centering
    \includegraphics[height=3.5cm, keepaspectratio]{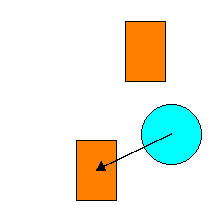}
\vskip -0.1in
\caption{A sample image (left) with its masked variant (right).}
\vskip -0.15in
\label{fig:syn_data}
\end{figure}

We construct a custom synthetic 2D benchmark, as shown in Fig \ref{fig:syn_data}, to analyze how VLM attention encodes entities and relations. The dataset consists of 5424 images in total, and is constructed using simple geometric shapes arranged in basic layouts with explicit ground-truth relations to aid mechanistic inspection. This design choice reduces linguistic ambiguity and enables controlled complexity.

\subsubsection{Scene Generator}
Each RGB image is 224$\times$224 canvas containing between 2--5 entities \{circle, square, rectangle, triangle\} with sampled attributes from 8 discrete colors, and 3 discrete sizes (small/medium/large). The entities are placed using rejection sampling with a fixed margin from borders and collision checks to avoid overlaps, ensuring clean layouts suitable for precise attention analysis. 

\subsubsection{Entity Relations}
For every scene, we compute pairwise spatial relations based on entity centroids and assign a primary axis-aligned relation from \{left\_of, right\_of, above, below\}. Furthermore, with a 50/50 split, we create \textbf{explicit} (arrow rendered on canvas) and \textbf{implicit} (no arrow; position-only) relationships.

\subsubsection{Masked Variants}\label{masked-vars}
To support causal tests of visual reliance, we generate a masked counterpart for each original image by re-rendering the scene while masking exactly one randomly chosen element: either (i) \textbf{entity} (remove one entity and its incident relations) or (ii) \textbf{relation} (mask one relation; if explicit, the arrow is omitted). Masking is implemented via full re-rendering rather than patching to avoid visual artifacts and to retain original scene information of entities and relations. These images constitute half of the dataset.

\subsection{Prompt Generation}
Complementarily, a set of prompts is generated for each image in the dataset. All prompts follow a multiple-choice format in order to aid quantitative analysis. For example:

\begin{quote}
\vskip -0.1in
\textbf{Query:} How many shapes are in the image? \\
\textbf{Options:} (a)\hspace{0.5em}4 \quad (b)\hspace{0.5em}0 \quad (c)\hspace{0.5em}2 \quad (d)\hspace{0.5em}5 \\
\textit{Instruction:} Please only reply with the correct option. If no option is correct, reply with 'None'.
\end{quote}
\vskip -0.1in
The queries are categorised based on their intended target of anlysis, and are further broken down into subtypes as shown in Table \ref{tab:queries}. They are generated using information contained within the image (specified in its corresponding JSON), which ensures a perfect mapping of queries to image content. For incorrect options provided in the prompt, existing entities and relations within the image are used - if not present or sufficient, then random entities and arrow connections contained in the dataset at large are used. 

\subsection{Attention Analysis}
Final model outputs and prediction accuracies are severely limited and insufficient to draw conclusions about model behavior. To this end, we employ \textbf{Linear Classifier Probes} \cite{alain2018understandingintermediatelayersusing}. It is a conceptual tool to better understand the dynamics inside a neural network and the role played by the intermediate layers. 

In this work, the linear probes act as layer-wise diagnostic for what information is readily decodable from Qwen-VL-4B's internal representation. Concretely, we extract visual attention features from each layer and train a lightweight linear classifier to predict query answers from a controlled synthetic benchmark of geometric shapes.

This analysis helps us characterize \textbf{emergence profiles}: which patterns appear early, which require deeper computation, and which fail to materialize. However, probe performance alone does not establish that the model causally relies on visual evidence to answer. To test reliance, we thus complemented probing with masked input variants (as outlined in Section \ref{masked-vars}) to track how accuracy is impacted. 

\section{Experiments and Discussion}

\subsection{Overall Query Performance} 
 \begin{table}[t]
\centering
\vskip 0.1in
\begin{small}
\begin{tabularx}{\columnwidth}{l l c}
\toprule
\textbf{Analysis Target} & \textbf{Query Type} & \textbf{Accuracy} (\%) \\
\midrule
\multirow{4}{*}{\textbf{Recognition}}
 & Count & 96.7 \\
 & Shape & 80.6 \\
 & Color & 96.8 \\
 & Shape \& Color & 99.5 \\
\midrule
\multirow{3}{*}{\textbf{Relational}}
 & Arrow Existence & 51.2 \\
 & Arrow Direction & 92.1 \\
 & Implicit Position & 61.8 \\
\bottomrule
\end{tabularx}
\end{small}
\vskip -0.1in
\caption{Accuracies for query types.}
\label{tab:acc}
\end{table}

The overall performance of the model is quantitatively measured by parsing model output and comparing against ground-truth dataset label. The accuracies achieved by the Qwen3-VL-4B model are summarized in Table \ref{tab:acc}. As expected, the model achieves high accuracy in recognition tasks. More specifically, the model performs consistently well when the query maps directly to tangible information in the visual input.

 We observe that \emph{Shape} yields much lower accuracies than \emph{Color}, pointing towards a much better color understanding of the VLM. Interestingly, \emph{Shape and Color} achieves an even higher accuracy, indicating performance improves with increased textual information. This becomes even more apparent when comparing the accuracies for the relational tasks: \emph{Arrow Direction} significantly outperforms \emph{Arrow Existence} and \emph{Implicit Position} queries. 
 
 This is due to the fact that more explicit information is given in the \emph{Arrow Direction} query, providing tangible and mappable information to visual data. Conversely, in \emph{Arrow Existence} queries, the VLM is tasked to explore all connections on its own without any textual clues in the query, and in \emph{Implicit Position} lack of tangible visual data leads to a significant accuracy drop.

\subsection{Linear Probing}
As the Qwen3-VL-4B is prompted, the attention maps are saved. These consist of $S\times L\times H$ where $S=5$ is the steps, $L=36$ is the number of layers, and $H=32$ the heads. The steps correspond to the most relevant attention steps, filtered to represent the 95th percentile of activity. To meaningfully extract information from them, we apply mean pooling across steps and heads. This leaves us with 36 attention maps per query. These are used to train simple layer-wise logistic regression models for each query, with 20\% of the data retained for testing. 

An additional caveat to note is that for simple queries of detection (color/shape/count), the probe is trained as is. However, for complex queries, the query was broken down into multiple sub-labels, and a separate linear probe had to be trained for each. The results discussed in the sections that follow are an average of performance over all sub-labels.
\begin{figure}
    \centering
    \includegraphics[width=1\linewidth]{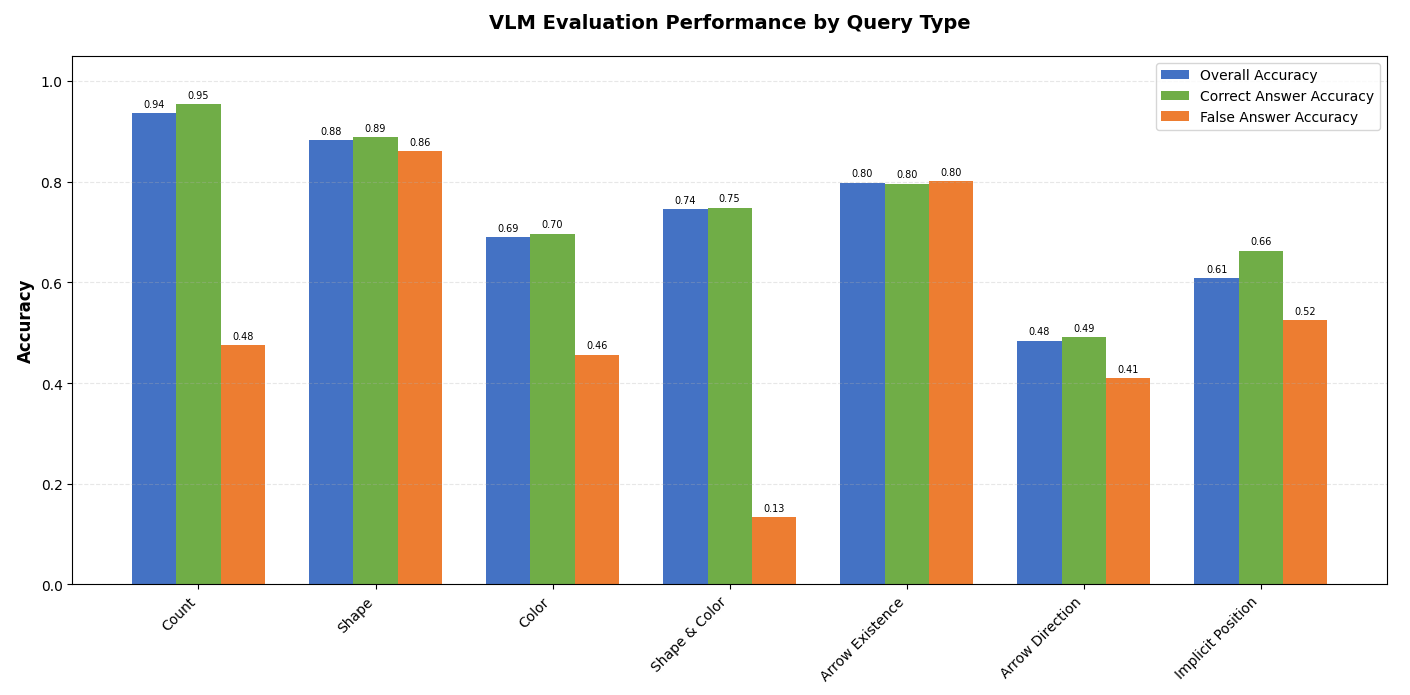}
    \vskip -0.2in
    \caption{VLM Performance from probing based on sample type.}
    \label{fig:probe_sample}
    \vskip -0.2in
\end{figure}

\begin{figure*}[t]
    \centering
    \includegraphics[width=\textwidth]{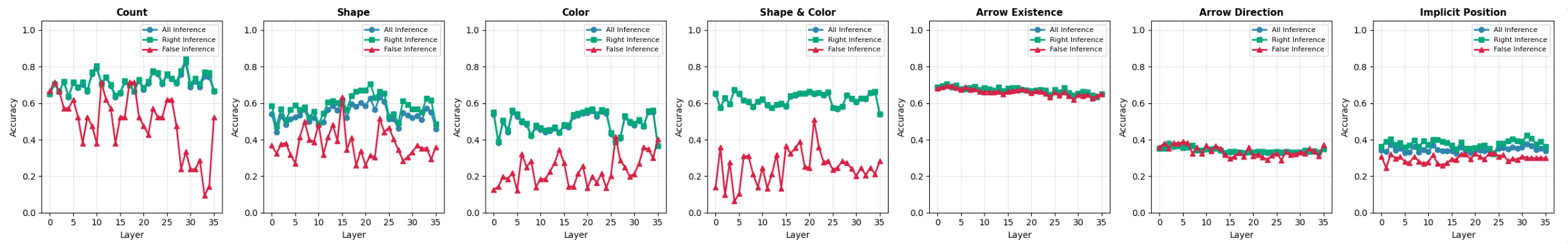}
    \vskip -0.15in
    \caption{Linear probing accuracies by query type across model layers.}
    \label{fig:probe_layers}
\end{figure*}
\subsubsection{Accuracies Based on Sample Types}
Figure \ref{fig:probe_sample} summarizes the probe accuracies clustered by sample types. These clusters correspond to the following:

\textbf{Overall Accuracy:} the overall decodability of that variable from the layer representation, computed on the entire evaluation set for a given query type.

\textbf{Correct Accuracy:} how well the representation supports correctly identifying positive cases, computed on the subset where the ground-truth answer is a positive class. 

\textbf{False Accuracy:} how well the representation supports correctly rejecting negatives, computed on the subset of samples where the ground-truth answer is the negative class.

\subsubsection{Accuracies Based on Query Type}
Figure \ref{fig:probe_layers} showcases accuracies across layers for the trained logistic models. We note the following findings: 

\textbf{Count:} Accuracy is consistently high and slightly increases toward mid–late layers, indicating that count information becomes more linearly decodable deeper in the network.

\textbf{Shape:} mid–late layers yield the highest accuracy on positive cases, while negative-cases remain lower and more variable. Interestingly, shape presence is more reliably encoded than color presence, i.e, geometric cues are easier to extract from attention representations than chromatic ones.

\textbf{Color:} shows moderate decodability with a clear asymmetry between positives and negatives. This indicates that attention features carry usable evidence for confirming a queried color, but provide comparatively weak evidence for rejecting absent colors.

\textbf{Shape and Color:} conjunction task amplifies the asymmetry: positive-case accuracy is relatively high and stable, whereas negative-case accuracy is consistently low. This is consistent with limited compositional binding - representation can confirm salient matching evidence, but struggles to confidently reject when the exact shape-color pair is absent.

\textbf{Arrow Existence:} this is comparatively well decoded and stable across depth, and shows near-overlap for sample types. This implies that the underlying relational signal (e.g., connection structure) is encoded consistently regardless of whether the model’s final answer is correct, pointing to robust representation-level encoding for this query type.

\textbf{Arrow Direction:} close to chance and unvaried across layers, with minimal separation between sample type. The attention-derived features do not provide a strong signal to resolve arrow-direction queries, and the model’s errors are not strongly concentrated in any specific subset.

\textbf{Implicit Position:} exhibits the weakest separability: sample types are tightly clustered, implying the representation contains only weak linearly decodable spatial information and, crucially, that the features are not markedly different between cases where the model succeeds or fails. In other words, spatial errors appear less driven by missing linearly accessible evidence and more by ambiguity/noise or non-linear dependencies not captured by the probe.

\subsubsection{Attention Patterns}
The following generalizations can be concluded from our experiments: \textbf{(i)} Early layers encode salient visual evidence and coarse statistics (often enough for presence and sometimes count), but weak for query identity and role-specific object identity. \textbf{(ii)} Middle layers have strongest regime for task-relevant visual signals (presence, counts, coarse relations), with the clearest separability for many probes. \textbf{(iii)} Late layers improve some higher-level relational structure (e.g., spatial position), but often do not recover object identity / compositional binding; some global signals (e.g., connection count) can even become less linearly accessible.

\subsection{Ablation Study}

To test whether high accuracy reflects genuine visual reasoning, we conducted the same experiments as outlined above for on the masked variant of the dataset as outlined in Section \ref{masked-vars}. The results are summarized in Table \ref{tab:ablation_results}. 

\begin{table}[t]
\centering
\label{tab:ablation_results}
\vskip 0.1in
\small
\begin{tabular}{lccc}
\toprule
\textbf{Query Type} & \textbf{Unmasked} & \textbf{Masked} & \textbf{Drop} \\
\midrule
Count & 96.7 & 74.1 & $-23.4$ \\
Arrow Direction & 92.1 & 76.4 & $-17.0$ \\
Shape \& Color & 99.5 & 99.3 & $-0.2$ \\
Color & 96.8 & 97.4 & $+0.6$ \\
Arrow Existence & 51.2 & 50.3 & $-1.7$ \\
Shape & 80.6 & 82.5 & $+2.4$ \\
Implicit Position & 61.8 & 64.5 & $+4.3$ \\
\bottomrule
\end{tabular}
\vskip -0.1in
\caption{Ablation study results (in \%) across all query types.}
\vskip -0.1in
\end{table}

Large drops such as those witnessed in \emph{Count} and \emph{Arrow Direction} do \textit{not} indicate successful visual grounding. Instead, they reveal hybrid strategies combining vision with internal biases. When masking creates distribution shift, these biases interfere with perception, causing hallucination. Other \emph{Recognition} tasks show changes within noise margins ($-0.2\%$ to $+2.4\%$). Surprisingly, \emph{Implicit Position} improves by $+4.3\%$---directly contradicting visual reasoning. This strongly indicates answers are derived from internal statistical regularities rather than visual content.

Our ablation exposes a dual-process architecture: hybrid tasks (count, arrows) attempt visual reasoning but fail under distribution shift ($-17\%$ to $-23\%$ drops), while shortcut tasks (recognition, spatial) bypass vision entirely ($\pm 5\%$ changes). High unmasked accuracy ($80$--$99\%$) creates an illusion of understanding that masking exposes as fragile.

\section{Conclusion}
Our work confirms that VLMs do not exhibit true visual \emph{understanding}. Their comprehension of visual information is strictly limited to what is explicitly observable, with no capacity to intelligently reason beyond it. The results show clear reliance on language cues and data priors - findings consistent with previous works. Our decoded attention patterns map this behavior to the models architecture, and opens up avenues for reformative approach and architecture changes to cater to this built-in limitation.

\bibliography{example_paper}

@misc{alain2018understandingintermediatelayersusing,
      title={Understanding intermediate layers using linear classifier probes}, 
      author={Guillaume Alain and Yoshua Bengio},
      year={2018},
      eprint={1610.01644},
      archivePrefix={arXiv},
      primaryClass={stat.ML},
      url={https://arxiv.org/abs/1610.01644}, 
}

@article{lu2023mathvista,
  title={Mathvista: Evaluating math reasoning in visual contexts with gpt-4v, bard, and other large multimodal models},
  author={Lu, Pan and Bansal, Hritik and Xia, Tony and Liu, Jiacheng and Li, Chunyuan and Hajishirzi, Hannaneh and Cheng, Hao and Chang, Kai-Wei and Galley, Michel and Gao, Jianfeng},
  journal={CoRR},
  year={2023}
}

@inproceedings{zhang2024mathverse,
  title={Mathverse: Does your multi-modal llm truly see the diagrams in visual math problems?},
  author={Zhang, Renrui and Jiang, Dongzhi and Zhang, Yichi and Lin, Haokun and Guo, Ziyu and Qiu, Pengshuo and Zhou, Aojun and Lu, Pan and Chang, Kai-Wei and Qiao, Yu and others},
  booktitle={European Conference on Computer Vision},
  pages={169--186},
  year={2024},
  organization={Springer}
}

@article{ainslie2023gqa,
  title={Gqa: Training generalized multi-query transformer models from multi-head checkpoints},
  author={Ainslie, Joshua and Lee-Thorp, James and De Jong, Michiel and Zemlyanskiy, Yury and Lebr{\'o}n, Federico and Sanghai, Sumit},
  journal={arXiv preprint arXiv:2305.13245},
  year={2023}
}

@article{clark2019does,
  title={What does bert look at? an analysis of bert's attention},
  author={Clark, Kevin and Khandelwal, Urvashi and Levy, Omer and Manning, Christopher D},
  journal={arXiv preprint arXiv:1906.04341},
  year={2019}
}

@article{elhage2021mathematical,
  title={A mathematical framework for transformer circuits},
  author={Elhage, Nelson and Nanda, Neel and Olsson, Catherine and Henighan, Tom and Joseph, Nicholas and Mann, Ben and Askell, Amanda and Bai, Yuntao and Chen, Anna and Conerly, Tom and others},
  journal={Transformer Circuits Thread},
  volume={1},
  number={1},
  pages={12},
  year={2021}
}

@article{hou2024vision,
  title={Do vision-language models really understand visual language?},
  author={Hou, Yifan and Giledereli, Buse and Tu, Yilei and Sachan, Mrinmaya},
  journal={arXiv preprint arXiv:2410.00193},
  year={2024}
}

@inproceedings{johnson2017clevr,
  title={Clevr: A diagnostic dataset for compositional language and elementary visual reasoning},
  author={Johnson, Justin and Hariharan, Bharath and Van Der Maaten, Laurens and Fei-Fei, Li and Lawrence Zitnick, C and Girshick, Ross},
  booktitle={Proceedings of the IEEE conference on computer vision and pattern recognition},
  pages={2901--2910},
  year={2017}
}

@inproceedings{masry2022chartqa,
  title={Chartqa: A benchmark for question answering about charts with visual and logical reasoning},
  author={Masry, Ahmed and Do, Xuan Long and Tan, Jia Qing and Joty, Shafiq and Hoque, Enamul},
  booktitle={Findings of the association for computational linguistics: ACL 2022},
  pages={2263--2279},
  year={2022}
}

@misc{bai2025qwen3vltechnicalreport,
      title={Qwen3-VL Technical Report}, 
      author={Shuai Bai and Yuxuan Cai and Ruizhe Chen and Keqin Chen and Xionghui Chen and Zesen Cheng and Lianghao Deng and Wei Ding and Chang Gao and Chunjiang Ge and Wenbin Ge and Zhifang Guo and Qidong Huang and Jie Huang and Fei Huang and Binyuan Hui and Shutong Jiang and Zhaohai Li and Mingsheng Li and Mei Li and Kaixin Li and Zicheng Lin and Junyang Lin and Xuejing Liu and Jiawei Liu and Chenglong Liu and Yang Liu and Dayiheng Liu and Shixuan Liu and Dunjie Lu and Ruilin Luo and Chenxu Lv and Rui Men and Lingchen Meng and Xuancheng Ren and Xingzhang Ren and Sibo Song and Yuchong Sun and Jun Tang and Jianhong Tu and Jianqiang Wan and Peng Wang and Pengfei Wang and Qiuyue Wang and Yuxuan Wang and Tianbao Xie and Yiheng Xu and Haiyang Xu and Jin Xu and Zhibo Yang and Mingkun Yang and Jianxin Yang and An Yang and Bowen Yu and Fei Zhang and Hang Zhang and Xi Zhang and Bo Zheng and Humen Zhong and Jingren Zhou and Fan Zhou and Jing Zhou and Yuanzhi Zhu and Ke Zhu},
      year={2025},
      eprint={2511.21631},
      archivePrefix={arXiv},
      primaryClass={cs.CV},
      url={https://arxiv.org/abs/2511.21631}, 
}

@misc{liu2024improvedbaselinesvisualinstruction,
      title={Improved Baselines with Visual Instruction Tuning}, 
      author={Haotian Liu and Chunyuan Li and Yuheng Li and Yong Jae Lee},
      year={2024},
      eprint={2310.03744},
      archivePrefix={arXiv},
      primaryClass={cs.CV},
      url={https://arxiv.org/abs/2310.03744}, 
}

@misc{wu2024gpt4visionhumanalignedevaluatortextto3d,
      title={GPT-4V(ision) is a Human-Aligned Evaluator for Text-to-3D Generation}, 
      author={Tong Wu and Guandao Yang and Zhibing Li and Kai Zhang and Ziwei Liu and Leonidas Guibas and Dahua Lin and Gordon Wetzstein},
      year={2024},
      eprint={2401.04092},
      archivePrefix={arXiv},
      primaryClass={cs.CV},
      url={https://arxiv.org/abs/2401.04092}, 
}

@misc{geminiteam2025geminifamilyhighlycapable,
      title={Gemini: A Family of Highly Capable Multimodal Models}, 
      author={Gemini Team Google},
      year={2025},
      eprint={2312.11805},
      archivePrefix={arXiv},
      primaryClass={cs.CL},
      url={https://arxiv.org/abs/2312.11805}, 
}
\bibliographystyle{icml2024}

\end{document}